**Dictionary-Learning-Based Reconstruction Method for Electron Tomography**

Baodong Liu[1,2], Hengyong Yu[1], Scott S. Verbridge[3], Lizhi Sun[4], Ge Wang[5]

1. Biomedical Imaging Division, VT-WFU School of Biomedical Engineering and Sciences, Wake Forest University Health Sciences, Winston-Salem, NC 27157
2. Key Lab of Nuclear Analysis Techniques, Institute of High Energy Physics, CAS; and Beijing Engineering Research Center of Radiographic Techniques and Equipment, Beijing 100049, PRC
3. Biomedical Imaging Division, VT-WFU School of Biomedical Engineering and Sciences, Virginia Tech, Blacksburg, VA 24060
4. Departments of Civil & Environmental Engineering and Chemical Engineering & Materials Science, University of California, Irvine, CA 92697
5. Biomedical Imaging Cluster, Department of Biomedical Engineering, Rensselaer Polytechnic Institute, Troy, NY 12180

*Correspondence to: Hengyong Yu (Hengyong-yu@ieee.org) or Ge Wang (ge-wang@ieee.org).*

**Key words.** Electron tomography, iterative reconstruction algorithm, simultaneous algebraic reconstruction technique (SART), linogram method, dictionary learning, sparse representation.

**Abstract**

Electron tomography usually suffers from so called "missing wedge" artifacts caused by limited tilt angle range. An equally sloped tomography (EST) acquisition scheme (which should be called the linogram sampling scheme) was recently applied to achieve 2.4-angstrom resolution. On the other hand, a compressive sensing-inspired reconstruction algorithm, known as adaptive dictionary based statistical iterative reconstruction (ADSIR), has been reported for x-ray computed tomography. In this paper, we evaluate the EST, ADSIR and an ordered-subset simultaneous algebraic reconstruction technique (OS-SART), and compare the ES and equally angled (EA) data acquisition modes. Our results show that OS-SART is comparable to EST, and the ADSIR outperforms EST and OS-SART. Furthermore, the equally sloped projection data acquisition mode has no advantage over the conventional equally angled mode in the context.

**I. Introduction**

Electron tomography (ET) targets ultra-fine details such as sub-cellular and macromolecular features. It uses a transmission electron microscope to collect data, and has generated a number of important results (Arslan, Yates et al. 2005, Lucic, Forster et al. 2005, Al-Amoudi, Diez et al. 2007, Robinson, Sali et al. 2007, Ben-Harush, Maimon et al. 2010). Nanometer ET resolution is commonly achieved. Conventional ET reconstructs a 3D object from a set of equally angled 2D projections. To avoid the interpolation between polar coordinates and Cartesian coordinates, Miao *et al* (Miao, Forster et al. 2005) developed an equally sloped tomography (EST), which made use of a set of equally sloped (ES) projections. In 2008, their team (Lee, Fahimian et al. 2008) developed an EST reconstruction algorithm to reconstruct the image from ES projections. The algorithm iterates back and forth between Fourier and object spaces. In each iteration, the calculated slices are updated with the experimentally measured slices in Fourier space and the physical constraints are enforced in the object space. Recently, Scott *et al.* (Scott, Chen et al. 2012) reported ET (using EST method) at 2.4-angstrom resolution as "*the experimental demonstration of a general ET method that achieves atomic-scale resolution without initial assumptions about the sample structure*".

Electron tomography is quite similar to x-ray computed tomography (CT), which reconstructs an image from line integrals. Great efforts have been made towards development of x-ray CT methods in the cases of incomplete and inaccurate data, especially truncated, limited angular and few-view imaging geometries. These reconstruction algorithms are generally in the state-of-the-art compressive sensing (CS) framework, utilizing prior knowledge effectively and permitting accurate and stable reconstruction from a more limited amount of raw data than requested by the classic Shannon sampling theory. CS-inspired reconstruction algorithms can be roughly categorized into the following generations (Wang, Bresler et al. 2011): (1) The 1st generation: Candes' total variation (TV) minimization method and variants (initially used for MRI and later on tried out for CT) (Li and Santosa 1996, Jonsson, Huang et al. 1998, Candes and Tao 2005, Landi and Piccolomini 2005, Yu, Li et al. 2005, Candes, Romberg et al. 2006, Block, Uecker et al. 2007, Candes, Wakin et al. 2008, Landi, Piccolomini et al. 2008, Sidky and Pan 2008, Yu and Wang 2009); (2) the 2nd generation: Soft-thresholding method adapted for x-ray CT to guarantee the convergence (Daubechies, Defrise et al. 2004, Yu and Wang 2010, Liu, Wang et al. 2011, Yu, Ji et al. 2011); and (3) the 3rd generation: Dictionary learning (DL) and non-local mean methods being actively developed by our group and others (Kreutz-Delgado, Murray et al. 2003, Gao, Yu et al. 2011, Bo, Huanjun et al. 2012, Lu, Zhao et al. 2012, Xu, Yu et al. 2012, Zhao, Ding et al. 2012).

For the 2nd generation algorithm, a pseudo inverse of the discrete difference transform was constructed with a soft-thresholding technique to perform the L1 minimization of the total difference. This method can be directly applied for few-view CT reconstruction, and showed superior performance over the 1st generation TV method.

As far as the 3rd generation reconstruction is concerned, dictionary learning has proven to be effective for extraction of sparsity. Recently, we combined dictionary learning and statistical reconstruction for few-view low-dose x-ray CT (Xu, Yu et al. 2012), in which a sparse constraint in terms of a redundant dictionary is incorporated into an objective function. The dictionary can be either pre-determined before an image reconstruction task or adaptively defined during the reconstruction process. Then, an alternating minimization algorithm is developed to minimize the objective function in a statistical iterative reconstruction framework. Our approach has been evaluated with low-dose x-ray projections collected in animal and human CT studies. Our results demonstrate that the DL approach can produce better images than the filtered backprojection (FBP) and TV-minimization algorithms (Xu, Yu et al. 2012).

In this study, we focus on the feasibility and merit of the dictionary learning approach for ET. We compare the EST, OS-SART and ADSIR in the ES data acquisition mode required by the EST reconstruction. We also compare the ES and equally angled (EA) data acquisition modes.

The rest of this paper is organized as follows. In section II, the EST method reported by Scott *et al*. (Scott, Chen et al. 2012), the classic ordered-subset simultaneous algebraic reconstruction technique (OS-SART), and an adaptive dictionary learning approach are briefly described for the purpose of ET reconstruction. Section III provides numerical comparative studies, followed by a discussion of relevant issues in Section IV. Finally, concluding remarks are given.

## II. Method

### 2.1. Equally sloped tomography

Electron tomography provides projection images of the specimen being imaged. A beam of electrons is shot towards the specimen, and scattered and unscattered electrons emerging from the specimen are then collected by magnetic lenses and focused to form an interference pattern, which constitutes the projection image (Frank 1992). In the ideal imaging condition, the projection images are formed by the integration of the 3D information of the specimen along the direction of the electron beam. The data acquisition process of ET can be modeled as the following linear system (Frank 2006)

$$p = Wf ,\qquad(1)$$

where $p$ represents projections, $f$ represents an electron image, $W$ is a measurement matrix. The image reconstruction is to solve $f$ from data $p$ for a given system matrix $W$. Because ET works in parallel-beam geometry, we can reconstruct a 3D image volume slice by slice.

In the noise free case, when the parallel projections are collected from a 180-degree coverage, the system (1) is well-posed and a good image can be easily reconstructed. However, the ET usually suffers from the so called limited angle problem, typically over a tilt range of +/- 60 or 70 degrees in small increments, which causes the "missing wedge" artifacts (Frank 1992).

In EST, the pseudopolar fast Fourier transform (PPFFT) and its inversion/adjoint algorithms are used to perform a fast Fourier transform for an object on a Cartesian grid. As a result, a Fourier slice is on a pseudopolar grid for tomographic reconstruction. To accommodate PPFFT and inversion/adjoint PPFFT, equally sloped projections are acquired by changing the angle with equal slop increments.

In the published experiments (Scott, Chen et al. 2012), the tilt angles ($\theta$) were determined by

$$\theta = \begin{cases} -\tan^{-1}[(N+2-2n)/N], (n=1,2,...,N) \\ \pi/2 - \tan^{-1}[(3N+2-2n)/N], (n=N+1, N+2,...,2N) \end{cases} \qquad(2)$$

with $N=32$ or 64 for a tilt range of $\pm 72.6^{\circ}$. For more detailed EST reconstruction method, please refer to (Miao, Forster et al. 2005, Lee, Fahimian et al. 2008, Scott, Chen et al. 2012).

## 2.2. Ordered-subset simultaneous algebraic reconstruction

When projection data are incomplete or noisy, the iterative reconstruction method is desirable to generate higher quality results than the filtered backprojection algorithm. Advanced iterative methods attract increasingly more attention because of their robustness against the experimental conditions in ET (Fernandez 2012). OS-SART (Wang and Jiang 2004) is a widely-used iterative reconstruction algorithm to solve Eq. (1) with guaranteed convergence (Wang and Jiang 2004).

When an algebraic reconstruction method is employed, the image $f$ can be discretized as a vector $f = (f_j) \in \mathbf{R}^{J \times 1}$ $(1 \leq j \leq J)$, where $J$ is the number of pixels. Accordingly, $p = (p_i) \in \mathbf{R}^{I \times 1}$ is a vector as well, where $p_i$ is the ray-sum associated with the $i^{th}$ ray and $I$ is the number of rays. Then, we have

$$p_i = \langle W_i, f \rangle = \sum_{j=1}^{J} w_{i,j} f_j, \quad i = 1, 2, ..., I, \tag{3}$$

where $W_i$ represents the $i^{th}$ row of $\mathbf{W}$, $w_{i,j}$ is the contribution of the $j^{th}$ pixel to the $i^{th}$ ray-sum. The pseudo-codes for the OS-SART are as follow:

*Initialization:* $k = 0$; Initialize $\hat{f}^0$ with an estimate image; Suppose that there are $N_{view}$ views. $\{T_0, T_1, ..., T_{L-1}\}$ is a partition of the set of views $\{1, 2, ..., N_{view}\}$, where $L$ is the number of subsets of views.
*While stopping criteria are not satisfied*
  *Update the current image*

$$\hat{f}_j^{k+1} = \hat{f}_j^k + \sum_{i \in T_m} \frac{w_{i,j}}{\sum_{n \in T_m} w_{n,j}} \frac{p_i - \langle W_i, \hat{f}^k \rangle}{\sum_{l=1}^{J} w_{i,l}}, \tag{4}$$

where $m = k \bmod L \in \{0, 1, ..., L-1\}$, $i \in T_m$ means the $i^{th}$ ray in $T_m$.

## 2.3. Dictionary learning-based reconstruction

Dictionary learning is effective for sparse representation. Recently, Xu *et al.* (Xu, Yu et al. 2012) developed global dictionary-based statistical iterative reconstruction (GDSIR) and adaptive dictionary-based statistical iterative reconstruction (ADSIR) for low-dose CT. The GDSIR requires a set of training images, which are often times not available. This is particularly the case of ET when samples to be studied are unknown. Hence, in this study we focus on ADSIR as an example of a highly versatile implementation of dictionary learning.

Let a vector $\hat{f} \in \mathbf{R}^{J \times 1}$ represent an image of $J_H \times J_W = J$ pixels. A dictionary is a matrix $\mathbf{D} \in \mathbf{R}^{N \times K}$ $(N \ll K)$ whose column $d_k \in \mathbf{R}^{N \times 1}$ $(k = 1, 2, ..., K)$ is called an atom. Further, $\mathbf{E}_s = \{e_{n,j}^s\} \in \mathbf{R}^{N \times J}$ $(s = 1, 2, ..., S)$ is a matrix to extract a $\sqrt{N} \times \sqrt{N}$ patch from the image $\hat{f}$, and $S = (J_H - \sqrt{N} + 1)(J_W - \sqrt{N} + 1)$ is the number of patches in a training set. A patch $\mathbf{E}_s \hat{f}$ is expected to be exactly or approximately represented as a sparse linear combination of the atoms in the dictionary $\mathbf{D}$; that is

$$\left\| \mathbf{E}_s \hat{f} - \mathbf{D}\alpha_s \right\|_2^2 \leq \varepsilon, \tag{5}$$

where $\varepsilon \geq 0$ is a small error bound, and the representation vector $\alpha_s \in \mathbf{R}^{K \times 1}$ has few nonzero entries, e.g., $\|\alpha_s\| \ll N \ll K$ with $\|\cdot\|_0$ being the $l_0$-norm.

The image reconstruction process using ADSIR is equivalent to solving the following optimization problem (Xu, Yu et al. 2012):

$$\min_{\hat{f}, \mathbf{D}, \alpha} \sum_{i=1}^{I} \frac{\tau_i}{2} \left( \langle \mathbf{W}_i, \hat{f} \rangle - p_i \right)^2 + \lambda \left( \sum_{s=1}^{S} \left\| \mathbf{E}_s \hat{f} - \mathbf{D}\alpha_s \right\|_2^2 + \sum_{s=1}^{S} v_s \|\alpha_s\|_0 \right), \tag{6}$$

where $\tau_i$ is the statistical weight for the $i^{th}$ x-ray path, $\lambda$ is a regularization parameter, $v = \{v_s\}_{s=1}^{S}$ is a Lagrange multiplier, and $\alpha \in \mathbf{R}^{K \times S}$ with $\alpha_s \in \mathbf{R}^{K \times 1}$. Although the statistical model is used to construct the likelihood to derive first term in Eq.(6) (Xu, Yu et al. 2012), it should be pointed out that the finally version of the first term of Eq.(6) can be viewed as a position-dependent weighting version of least square for data discrepancy, which is a variant of the conventional SART reconstruction. Meanwhile, the first term of Eq.(6) can be simplified to the conventional data fidelity term $\frac{1}{2}\|W\hat{f} - p\|_2^2$ if we assume the detected photon numbers are same at all the detector cells.

To improve the performance of the ADSIR method (Xu, Yu et al. 2012), here we use the ordered-subset technique and update the dictionary every $N_{iterval}$ iterations. The detailed description of ADSIR and the selection of the parameters can be found in (Xu, Yu et al. 2012). The sparse level $L_0^s$ is the number of atoms involved in representing a patch, which is empirically determined according to the complexity of an image to be reconstructed and the property of the dictionary. The pseudo-codes for the new ADSIR are listed below.

*Choose $\lambda$, $\varepsilon$, $L_0^s$;*

*Initialize $\hat{f}^0$ (zero image or the reconstructed image by other algorithms), $\mathbf{D}^0$ (trained from the extracted sets from $\hat{f}^0$), $\alpha^0$, and $k = 0$; Suppose that there are $N_{view}$ views. $\{T_0, T_1, ..., T_{L-1}\}$ is a partition of a set of views $\{1, 2, ..., N_{view}\}$, $L$ is the number of subsets of views;*

*While the stopping criteria are not satisfied*
  *1) For $m = 0, 1, ..., L-1$*

$$\hat{f}_j^k = \hat{f}_j^k - \frac{\sum_{i \in T_m} \left( w_{i,j} \left( \langle \mathbf{W}_i, \hat{f}^k \rangle - p_i \right) \right) + 2\lambda \sum_{s=1}^{S} \sum_{n=1}^{N} e_{n,j}^s \left( \left[ \mathbf{E}_s \hat{f}^k \right]_n - \left[ \mathbf{D}^k \alpha_s^k \right]_n \right)}{\sum_{i \in T_m} \left( w_{i,j} \sum_{l=1}^{J} w_{i,l} \right) + 2\lambda \sum_{s=1}^{S} \sum_{n=1}^{N} e_{n,j}^s \sum_{l=1}^{J} e_{n,l}^s}, \quad j = 1, 2, ..., J; \tag{7}$$

  *2) $\hat{f}^{k+1} = \hat{f}^k$, $k = k+1$;*

3) If $k \bmod N_{iterval} = 0$

    *Extract patches from $\hat{f}^k$ to form a training set;*

    *Construct a dictionary $\mathbf{D}^k$ from the training set;*

  *Else*

    $\mathbf{D}^k = \mathbf{D}^{k-1}$;

4) *Represent $\hat{f}^k$ with a sparse $\alpha^k$ in terms of the dictionary $\mathbf{D}^k$ using the orthogonal matching pursuit (OMP) method;*

  *Output the final reconstruction.*

## III. Simulation results and discussions

In June, 2012, Dr. Miao, the corresponding author of (Scott, Chen et al. 2012), made available to us the following items used in (Scott, Chen et al. 2012):

- A 121x121x121 phantom of 0.5 angstrom voxel size (shown in Figure 1);
- 55 and 69 projections (-72.6° to 72.6° in an "equally sloped tomography" setting);
- Both loose and tight supports of the model;
- EST reconstructed results from 55 ES projections with loose support after 500 iterations.

Because the current study is to investigate electron tomography for atom-level imaging, the phantom provided by Dr. Miao is a most suitable choice to mimic atoms. The maximum value of the original phantom is $1.61 \times 10^5$, while the maximum value of the EST reconstruction from 55 projections is $2.79 \times 10^7$. This implies that the scale of Dr. Miao's EST results is inconsistent with the original phantom. During the course of the review process of this paper, we noticed that Dr. Miao's group has published the corresponding software on their group's webpage (*http://www.physics.ucla.edu/research/imaging/EST/index.htm*) and the scale problem has been corrected. Hence, we repeated their numerical simulations with updated results for comparison and analysis.

    To explore the effectiveness of the reconstruction algorithms with fewer views, assuming the same scanning range (-72.6° to 72.6°), 31 projections in an "equally sloped tomography" setting were also simulated. Therefore, the 3D phantom was reconstructed from 69/55/31 ES and equally angled (EA) projections using EST (not applicable to EA projections), OS-SART and ADSIR.

    In our experiments, each view was set to be a subset. The loose support was assumed for all the reconstruction methods. For the EST, the reconstruction parameters were set to be the same as those used in (Scott, Chen et al. 2012). The loose support was used in the first 500 iterations in the EST algorithm. Then, a tight support was determined based on the intermediate result. Using the tight support, the EST algorithm run another 500 iterations to obtain the final 3D image. For the OS-SART, 200 iterations were used. For the ADSIR, the result obtained by the OS-SART after 100 iterations was set as the initial image. The final result was obtained after 100 ADSIR iterations with $N_{iterval} = 10$. The parameters for the ADSIR were chosen as $\lambda = 0.1$, $\varepsilon = 5.0 \times 10^{-6}$, $L_0^S = 8$, $N = 64$ and $K = 256$. To ensure the redundancy, the number of atoms $K$ in a dictionary

should be much greater than that of pixels in a patch, which means $K \gg N$. In the image processing field, it has been proved that $K=4N$ is sufficient (Xu, Yu et al. 2012). If $N$ is too small, the atoms can not capture the typical features of the trained image. The greater the patch size $N$ is, the more the computational cost requires. In practical applications, $N=64$ is an optimized parameter.

To compare the results quantitatively, the results were quantitatively evaluated using two indices. One is the root mean square error (RMSE),

$$\text{RMSE}(f, f^*) = \sqrt{\frac{1}{J} \sum_{j=1}^{J} (f_j - f_j^*)^2}, \qquad (8)$$

where $f = (f_j) \in \mathbf{R}^{J \times 1}$ represents a reconstructed image, and $f^* = (f_j^*) \in \mathbf{R}^{J \times 1}$ is the reference phantom. The other is the image quality assessment index for structural similarity (SSIM) (Wang, Bovik et al. 2004), which is shown to be consistent with visual perception. The closer it is to 1, the higher the structural similarity is.

*3.1. Comparison of reconstruction methods*

The RMSE and SSIM (the averaged SSIM values for all slices) indices of EST, OS-SART and ADSIR results are listed in Table I. Figure 2 plots the profiles of RMSE and SSIM in Table I for comparison of the EST, OS-SART and ADSIR methods. From the above quantitative comparisons, we can see that the performance of OS-SART is comparable to the EST, and the ADSIR outperforms the EST and OS-SART especially when fewer projections were used. The ADSIR reconstructed higher quality images than the EST and OS-SART from 31 projections. When the number of views was reduced from 69 to 31, the reconstructed image quality by the ADSIR decreased much less than that by the EST and OS-SART. In other words, ADSIR is more robust than the EST and OS-SART.

Our codes were developed on a PC with i5 CPU 760 and a 4GB RAM. The average computational times were recorded for one slice and one loop of the OS-SART (0.61s) and the ADSIR (10.57s for dictionary learning, 15.15s for the other steps) with 55 projections. Although the computational cost of the ADSIR is high, it should not be a bottleneck for practical applications because graphics processing unit (GPU) and other hardware-based high-performance computing technologies are being developed.

*3.2 Comparison of data acquisition modes*

As we discussed in Subsection 2.1, the ES data acquisition mode is required by the EST reconstruction method. In this subsection, the ES and EA data acquisition modes will be compared in the context of OS-SART and ADSIR reconstruction methods.

Figure 3 plots the profiles of RMSE and SSIM in Table I for comparison of the ES and EA data acquisition modes. It can be seen that the results from the ES mode are very similar to the results from the EA mode when using OS-SART and ADSIR reconstruction methods. This implies that the ES data acquisition mode has no advantage over the EA data acquisition mode.

Indeed, it is well known that straight-ray tomography (such as x-ray CT and ET) could not benefit from ES sampling relative to EA sampling. The former is less even than the latter (see Figure 4). In the ES mode, because $N$ in formula (2) is usually selected as a power of 2 for fast Fourier transform, in most cases the samplings are uneven. For example, we can get 107 relatively even views located in a tilt range of $\pm 72.6º$ from formula (2) with $N=64$. In order to get 69 views, 67 uneven views were picked and the other two views at -72.6º and 72.6º were added. In fact, it is the EA sampling that is popular in practice (Natterer and Wubbeling 2001), even for missing wedge problems.

With proper constraints or prior knowledge, the missing wedge problem can be addressed to various degrees, which has been widely used for many years. This improvement is irrelevant to the ES strategy. Clearly, the EST method is a combination of the linogram method and some routine iterative reconstruction techniques (enhancing object support and positivity). As confirmed in (Averbuch, Coifman et al. 2008), "*Edholm and Herman contributed the fundamental insight that there was a continuum transform that could be discretized compatibly, provided one thinks in terms of slopes rather than angles.*" Excluding the well-known iterative enhancement of popular constraints, EST is essentially the same as the linogram method published two decades ago (Edholm and Herman 1987). Mathematical results were proved in (Edholm and Herman 1987) to appropriately parameterize the image and Fourier spaces for computational benefits.

## IV. Conclusions

While equally sloped sampling avoids some interpolation, its sampling geometry is not symmetric, being disadvantageous relative to equiangular sampling. As shown in this work, equally sloped sampling does not contribute to refining image quality nor alleviating the missing wedge problem. As confirmed in (Averbuch et al 2008), equally sloped sampling is the same as the linogram sampling method published before (Edholm and Herman, 1987).

In conclusion, we evaluated the EST with ADSIR and an OS-SART assuming the same data model used in (Scott, Chen et al. 2012). Our results show that (I) OS-SART is comparable to EST, and ADSIR outperforms EST and OS-SART; (II) the equally sloped projection data acquisition mode has no advantage over the conventional equally angled mode when using OS-SART and ADSIR algorithms. These results can provide a valuable reference in the field of straight-ray tomography including but not limited to ET and CT.

**Acknowledgements**
This work was partially supported by the NIH/NIBIB Grant EB011785, the NSF CAREER Award CBET-1149679, the NSF Collaborative project DMS-1210967, as well as the NSF grant CMMI-1229405. The work of Liu was partially supported by IHEP-CAS Scientific Research Foundation 2013IHEPYJRC801. The authors thank Dr. Miao for academic discussion and simulated data.

**References**

Al-Amoudi, A., D. C. Diez, et al. (2007). "The molecular architecture of cadherins in native epidermal desmosomes." <u>Nature</u> **450**(7171): 832-U838.

Arslan, I., T. J. V. Yates, et al. (2005). "Embedded nanostructures revealed in three dimensions." *Science* **309**(5744): 2195-2198.

Averbuch, A., R. R. Coifman, et al. (2008). "A framework for discrete integral transformations I - The pseudopolar Fourier transform." *Siam Journal on Scientific Computing* **30**(2): 764-784.

Ben-Harush, K., T. Maimon, et al. (2010). "Visualizing cellular processes at the molecular level by cryo-electron tomography." *Journal of Cell Science* **123**(1): 7-12.

Block, K. T., M. Uecker, et al. (2007). "Undersampled radial MRI with multiple coils. Iterative image reconstruction using a total variation constraint." *Magnetic Resonance in Medicine* **57**(6): 1086-1098.

Bo, Z., D. Huanjun, et al. (2012). "Dual-dictionary learning-based iterative image reconstruction for spectral computed tomography application." *Physics in Medicine and Biology* **57**(24): 8217.

Candes, E. J., J. Romberg, et al. (2006). "Robust uncertainty principles: exact signal reconstruction from highly incomplete frequency information." *IEEE Transactions on Information Theory* **52**(2): 489-509.

Candes, E. J. and T. Tao (2005). "Decoding by linear programming." *Information Theory, IEEE Transactions on* **51**(12): 4203-4215.

Candes, E. J., M. B. Wakin, et al. (2008). "Enhancing Sparsity by Reweighted l(1) Minimization." *Journal Of Fourier Analysis And Applications* **14**(5-6): 877-905.

Daubechies, I., M. Defrise, et al. (2004). "An iterative thresholding algorithm for linear inverse problems with a sparsity constraint." *Communications On Pure And Applied Mathematics* **57**(11): 1413-1457.

Edholm, P. R. and G. T. Herman (1987). "Linograms in image reconstruction from projections." *Ieee Transactions on Medical Imaging* **6**: 301-307.

Fernandez, J. J. (2012). "Computational methods for electron tomography." *Micron* **43**(10): 1010-1030.

Frank, J. (1992). *Electron tomography : three-dimensional imaging with the transmission electron microscope*. New York ; London, Plenum.

Frank, J. (2006). *Electron tomography: methods for three-dimensional visualization of structures in the cell*, Springer.

Gao, H., H. Y. Yu, et al. (2011). "Multi-energy CT based on a prior rank, intensity and sparsity model (PRISM)." *Inverse Problems* **27**(11).

Jonsson, E., S. C. Huang, et al. (1998). "Total variation regularization in positron emission tomography." *CAM report*: 98-48.

Kreutz-Delgado, K., J. F. Murray, et al. (2003). "Dictionary learning algorithms for sparse representation." *Neural Computation* **15**(2): 349-396.

Landi, G. and E. L. Piccolomini (2005). "A total variation regularization strategy in dynamic MRI." *Optimization Methods & Software* **20**(4-5): 545-558.

Landi, G., E. L. Piccolomini, et al. (2008). "A total variation-based reconstruction method for dynamic MRI." *Computational and Mathematical Methods in Medicine* **9**(1): 69-80.

Lee, E., B. P. Fahimian, et al. (2008). "Radiation dose reduction and image enhancement in biological imaging through equally-sloped tomography." *Journal of Structural Biology* **164**(2): 221-227.

Li, Y. Y. and F. Santosa (1996). "A computational algorithm for minimizing total variation in image restoration." *IEEE Transactions on Image Processing* **5**(6): 987-995.

Liu, B., G. Wang, et al. (2011). "Image reconstruction from limited angle projections collected by multisource interior x-ray imaging systems." *Physics in Medicine and Biology* **56**(19): 6337-6357.

Lu, Y., J. Zhao, et al. (2012). "Few-view image reconstruction with dual dictionaries." *Physics in Medicine and Biology* **57**(1): 173-189.

Lucic, V., F. Forster, et al. (2005). "Structural studies by electron tomography: From cells to molecules." *Annual Review of Biochemistry* **74**: 833-865.

Miao, J. W., F. Forster, et al. (2005). "Equally sloped tomography with oversampling reconstruction." *Physical Review B* **72**(5).

Natterer, F. and F. Wubbeling, Eds. (2001). *Mathematical Methods in Image Reconstruction*. Philadelphia, Society for Industrial and Applied Mathematics.

Robinson, C. V., A. Sali, et al. (2007). "The molecular sociology of the cell." *Nature* **450**(7172): 973-982.

Scott, M. C., C. C. Chen, et al. (2012). "Electron tomography at 2.4-angstrom resolution." *Nature* **483**(7390): 444-447.

Sidky, E. Y. and X. Pan (2008). "Image reconstruction in circular cone-beam computed tomography by constrained, total-variation minimization." *Physics in Medicine and Biology* **53**(17): 4777-4807.

Wang, G., Y. Bresler, et al. (2011). "Compressive Sensing for Biomedical Imaging." *Ieee Transactions on Medical Imaging* **30**(5): 1013-1016.

Wang, G. and M. Jiang (2004). "Ordered-subset simultaneous algebraic reconstruction techniques (OS-SART)." *Journal of X-Ray Science and Technology* **12**(3): 169-177.

Wang, Z., A. C. Bovik, et al. (2004). "Image quality assessment: From error visibility to structural similarity." *IEEE Transactions on Image Processing* **13**(4): 600-612.

Xu, Q., H. Yu, et al. (2012). "Low-dose X-ray CT Reconstruction via Dictionary Learning." *Medical Imaging, IEEE Transactions on* **31**(9): 1682-1697.

Yu, G. Q., L. Li, et al. (2005). "Total variation based iterative image reconstruction." *Computer Vision for Biomedical Image Applications, Proceedings* **3765**: 526-534.

Yu, H., C. Ji, et al. (2011). "SART-Type Image Reconstruction from Overlapped Projections." *Int J Biomed Imaging* **2011**.

Yu, H. and G. Wang (2010). "A soft-threshold filtering approach for reconstruction from a limited number of projections." *Physics in Medicine and Biology* **55**(13): 3905-3916.

Yu, H. Y. and G. Wang (2009). "Compressed sensing based interior tomography." *Physics in Medicine and Biology* **54**(9): 2791-2805.

Zhao, B., H. Ding, et al. (2012). "The Feasibility of the Dual-Dictionary Method for Breast Computed Tomography Based On Photon-Counting Detectors." *Medical Physics* **39**(6): 3915-3916.

Table I. Quantitative analysis on OS-SART and ADSIR.

| Views # | Reconstruction methods | Equally sloped acquisition mode | | Equally angled acquisition mode | |
|---|---|---|---|---|---|
| | | RMSE | SSIM | RMSE | SSIM |
| 69 | EST | 749.8 | 0.9935 | N/A | N/A |
| | OS-SART | 615.0 | 0.9923 | 608.6 | 0.9923 |
| | ADSIR | 386.8 | 0.9962 | 388.3 | 0.9960 |
| 55 | EST | 966.8 | 0.9905 | N/A | N/A |
| | OS-SART | 819.9 | 0.9883 | 833.8 | 0.9878 |
| | ADSIR | 395.5 | 0.9960 | 399.9 | 0.9960 |
| 31 | EST | 1840.4 | 0.9694 | N/A | N/A |
| | OS-SART | 1719.4 | 0.9627 | 1716.2 | 0.9629 |
| | ADSIR | 550.8 | 0.9936 | 564.9 | 0.9935 |

**Figures**

*Fig. 1. Original phantom images. While the top row shows a 3D view, the bottom row shows three central slices in a display window [0, $10^5$].*

*Fig. 2. Plots of RMSE and SSIM in Table I for comparison of the EST, OS-SART and ADSIR methods.*

*Fig. 3. Plots of RMSE and SSIM in Table I for comparison of the ES and EA data acquisition modes.*

*Fig. 4. Illustration of equally sloped and equally angled acquisition modes. The projection directions are marked with short lines.*

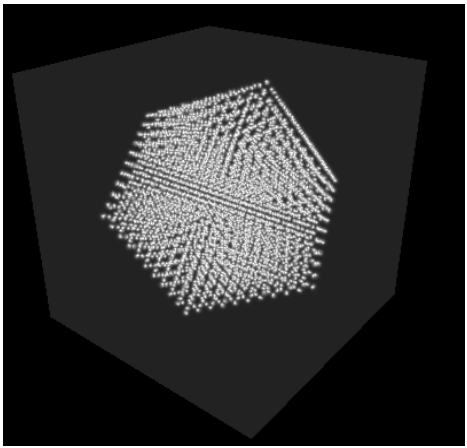
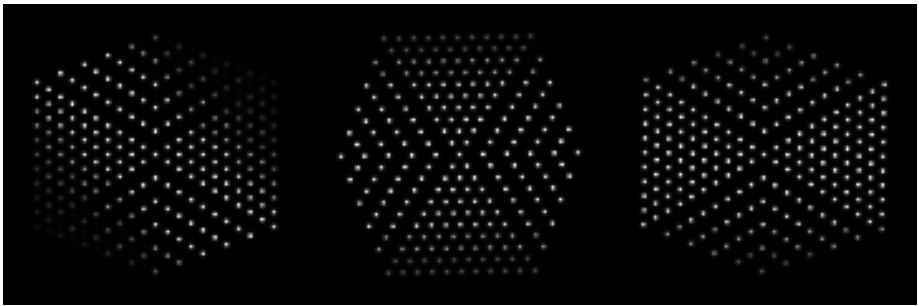

Fig. 1.

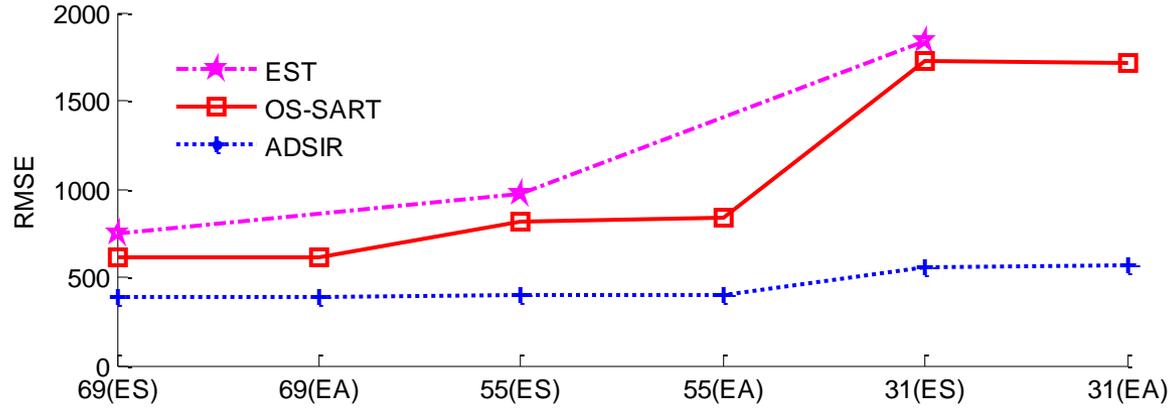
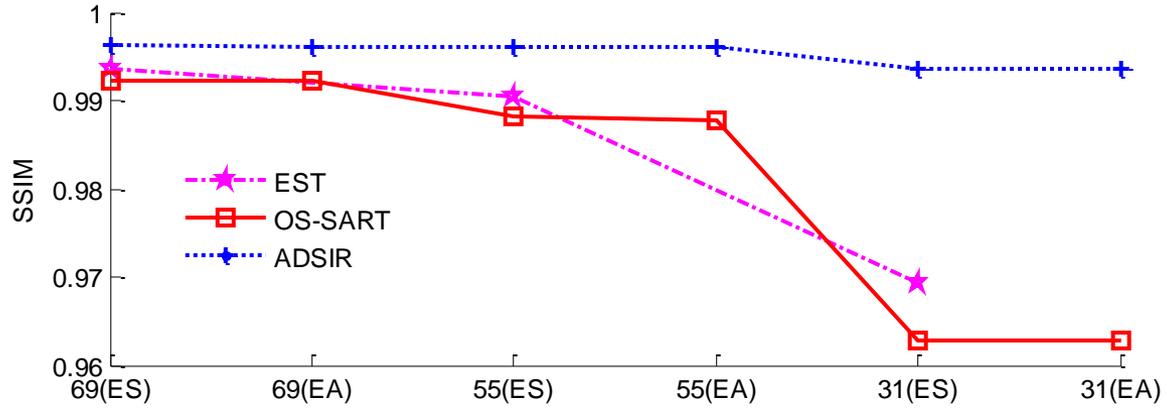

Fig. 2.

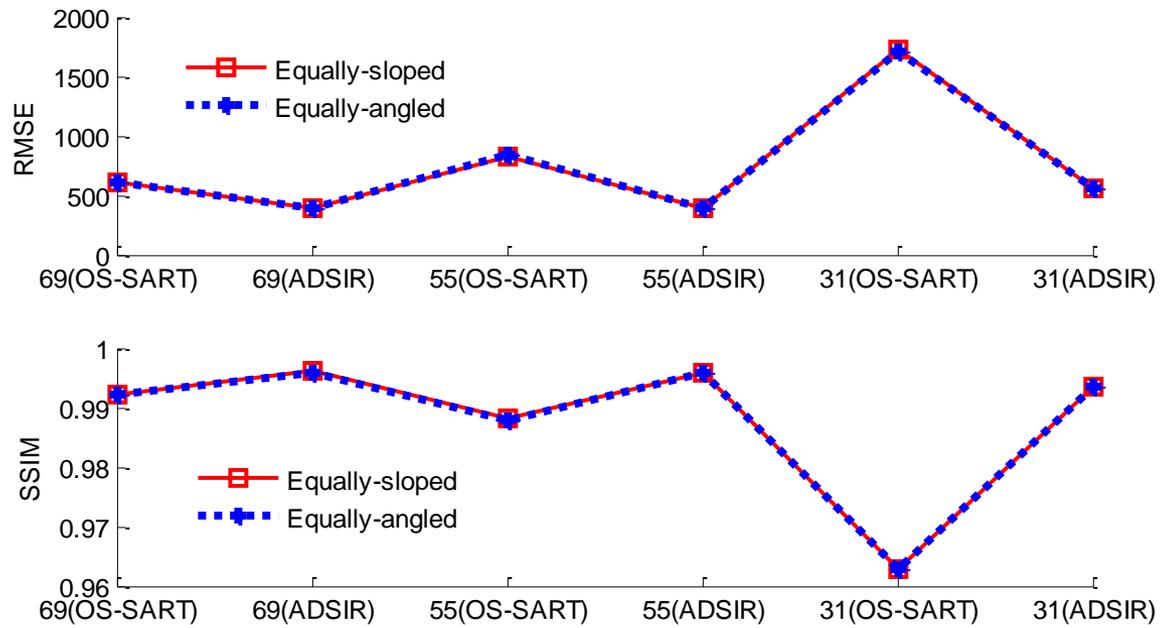

Fig. 3.

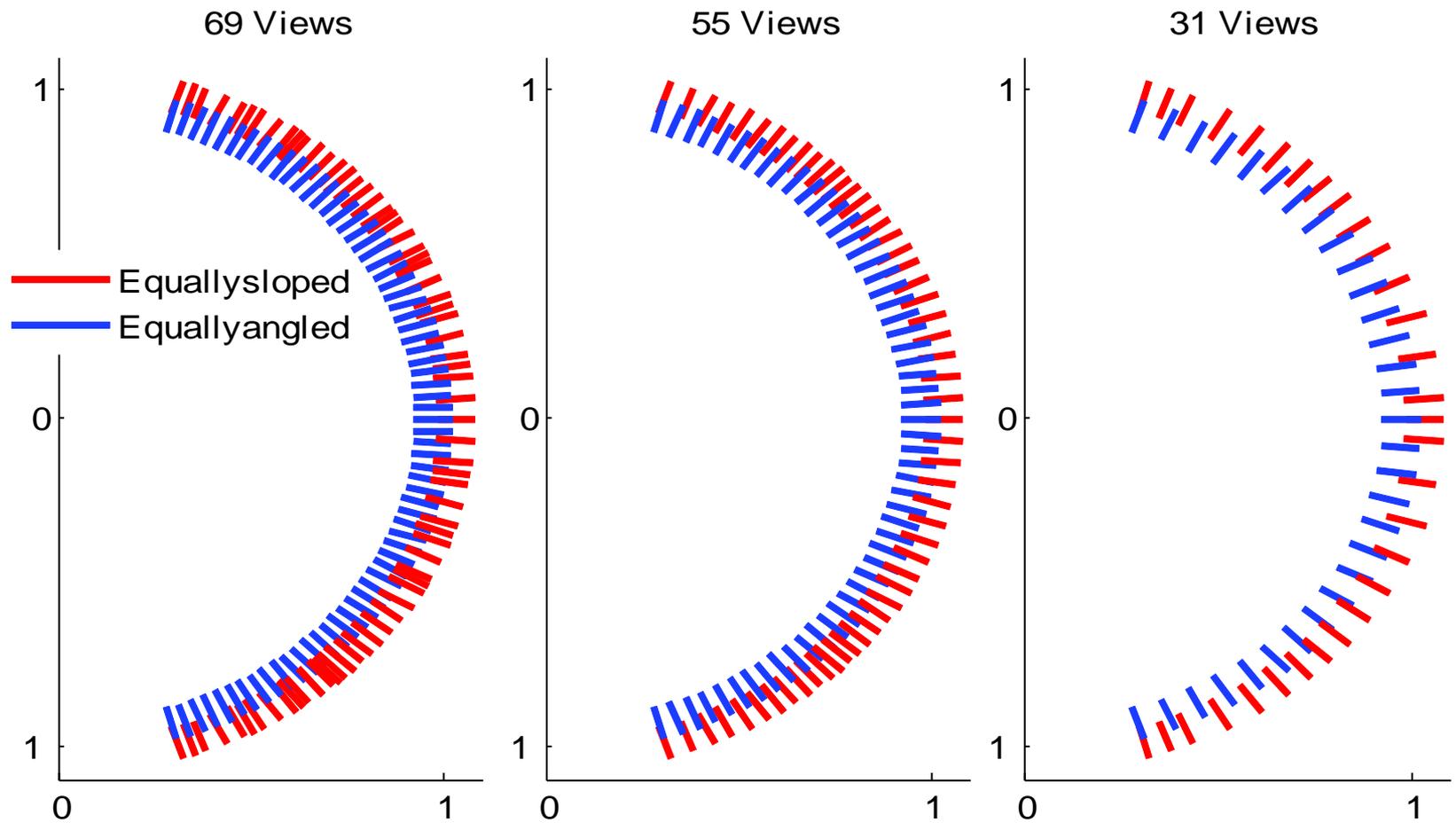

Fig. 4.